\documentclass{article}

\usepackage{PRIMEarxiv}

\usepackage[utf8]{inputenc} % allow utf-8 input
\usepackage[T1]{fontenc}    % use 8-bit T1 fonts
\usepackage{hyperref}       % hyperlinks
\usepackage{url}            % simple URL typesetting
\usepackage{booktabs}       % professional-quality tables
\usepackage{amsfonts}       % blackboard math symbols
\usepackage{nicefrac}       % compact symbols for 1/2, etc.
\usepackage{microtype}      % microtypography
\usepackage{lipsum}
\usepackage{fancyhdr}       % header
\usepackage{graphicx}       % graphics
\graphicspath{{media/}}     % organize your images and other figures under media/ folder

\usepackage{tabularx,booktabs}
\usepackage{wrapfig}
\usepackage[utf8]{inputenc} % allow utf-8 input
\usepackage[T1]{fontenc}    % use 8-bit T1 fonts
\usepackage{hyperref}       % hyperlinks
\usepackage{url}            % simple URL typesetting
\usepackage{booktabs}       % professional-quality tables
\usepackage{amsfonts}       % blackboard math symbols
\usepackage{nicefrac}       % compact symbols for 1/2, etc.
\usepackage{microtype}      % microtypography
\usepackage{lipsum}
\usepackage{fancyhdr}       % header
\usepackage{graphicx}       % graphics
\graphicspath{{media/}}     % organize your images and other figures under media/ folder
\usepackage{seqsplit}
\usepackage{geometry}

\usepackage{subcaption}
\usepackage{amsthm,amssymb,amsfonts}
\usepackage{tikz,lipsum,lmodern}
\usepackage[most]{tcolorbox}
\definecolor{cream}{RGB}{222,217,201}

\newtcolorbox{Box1}[2][]{
                lower separated=false,
                colback=white!80!gray,
colframe=black, fonttitle=\bfseries,
colbacktitle=black!50!gray,
coltitle=black,
enhanced,
attach boxed title to top left={xshift=0.5cm,yshift=-2mm},
title=#2,
boxrule=0.5pt,
boxed title style={colframe=black, boxrule=0.5pt},
#1}

\newtcolorbox{Box2}[2][]{
                lower separated=false,
                colback=white!80!white,
colframe=black, fonttitle=\bfseries,
colbacktitle=white!50!white,
coltitle=black,
enhanced,
attach boxed title to top center={yshift=-2mm},
title=#2,
boxrule=0.5pt,
boxed title style={colframe=black, boxrule=0.5pt},
#1,}

\title{AtomAgents: Alloy design and discovery through physics-aware multi-modal multi-agent artificial intelligence
\thanks{\textit{\underline{Citation}}: 
\textbf{A. Ghafarollahi, M.J. Buehler. arXiv, DOI:000000/11111., 2024}} 
}

\author{
  Alireza Ghafarollahi \\
  Laboratory for Atomistic and Molecular Mechanics (LAMM)\\Massachusetts Institute of Technology\\ 77 Massachusetts Ave.\\ Cambridge, MA 02139, USA 
   \And
  Markus J. Buehler \\
  Laboratory for Atomistic and Molecular Mechanics (LAMM)  \\
  Center for Computational Science and Engineering\\ Schwarzman College of Computing,\\ Massachusetts Institute of Technology\\77 Massachusetts Ave.\\Cambridge, MA 02139, USA\\ \\
  Correspondence: \texttt{mbuehler@MIT.EDU} \\
}

\begin{document}
\maketitle

\begin{abstract}
The design of new alloys is a multi-scale problem that requires a holistic approach that involves retrieving relevant knowledge, applying advanced computational methods, conducting experimental validations, and analyzing the results, a process that is typically slow and reserved for human experts. Machine learning (ML) can help accelerate this process, for instance, through the use of deep surrogate models that connect structural and chemical features to material properties, or \textit{vice versa}. However, existing data-driven models often target specific material objectives, offering limited flexibility to integrate out-of-domain knowledge and cannot adapt to new, unforeseen challenges. Here, we overcome these limitations by leveraging the distinct capabilities of multiple AI agents that collaborate autonomously within a dynamic environment to solve complex materials design tasks. The proposed physics-aware generative AI platform, AtomAgents, synergizes the intelligence of large language models (LLM) the dynamic collaboration among AI agents with expertise in various domains, including knowledge retrieval, multi-modal data integration, physics-based simulations, and comprehensive results analysis across modalities that includes numerical data and images of physical simulation results. The concerted effort of the multi-agent system allows for addressing complex materials design problems, as demonstrated by examples that include autonomously designing metallic alloys with enhanced properties compared to their pure counterparts. Our results enable accurate prediction of key characteristics across alloys and highlight the crucial role of solid solution alloying to steer the development of advanced metallic alloys. Our framework enhances the efficiency of complex multi-objective design tasks and opens new avenues in fields such as biomedical materials engineering, renewable energy, and environmental sustainability.
\end{abstract}

% keywords can be removed
\keywords{Multi-agent system \and Mechanics and materials \and Large language model \and Materials design \and Atomistic simulation \and Scientific machine learning \and Multi-modal generative AI \and Materials science \and Natural language processing}

\section{Introduction}
The continuous demand for new materials is driven by the need to address emerging technological challenges, enhance efficiencies, reduce costs, and minimize environmental impacts across a range of industries \cite{miracle2014exploration, wang2019advanced, raabe2023materials, shen2023computational}. Innovations in materials science can catalyze breakthroughs in sectors such as electronics \cite{rogers2010materials}, aerospace \cite{zhang2018recent, blakey2021metal}, energy storage \cite{liu2010advanced, koohi2020review}, and biomedicine \cite{lin2019muscle}. For example, the development of lighter, stronger materials could lead to more fuel-efficient vehicles and aircraft, while advancements in semiconductor technology could revolutionize electronics through enhanced functionality and reduced energy consumption~\cite{chaves2020bandgap}.

Metal alloys are indispensable for many structural applications mainly due to the critical role of defects in their crystalline lattices, such as dislocations, interfaces, crack tips, grain boundaries, precipitates, and vacancies. These defects and their interactions determine key properties like plastic flow behavior, creep, fatigue, and fracture toughness, directly impacting the material's performance \cite{suresh1998fatigue, buehler2008atomistic, wu2016mechanism, wu2018mechanistic, mak2021ductility, }. Understanding these defects and their chemical dependencies is crucial for optimizing existing alloys and designing new, high-performance materials. Experimental and computational methods are essential in this pursuit, providing detailed insights into the complex behaviors of metallic systems under varying conditions. Advanced experimental techniques reveal the formation and dynamics of defects in real-time, enhancing our understanding of material behavior under operational conditions. Atomistic simulations, including molecular dynamics (MD) \cite{ buehler2003hyperelasticity, buehler2004atomic, pellenq2009realistic, taheri2019transition} and Density Functional Theory (DFT) \cite{ventelon2013ab}, offer detailed models of defect behaviors at the atomic level, enabling the prediction of material behaviors under various conditions. These are complemented by physics-based theoretical models, which bridge the gap between atomic-scale phenomena and macroscopic material behavior, thereby further accelerating the material design process and providing mechanistic insights.~\cite{buehler2003atomistic, buehler2004hierarchical, hartmaier2005multiscale, ghafarollahi2020theory, ghafarollahi2021theory,kubilay2021high, ghafarollahi2022screw, maresca2020theory, varvenne2017solute} Recently, machine learning (ML) and artificial intelligence (AI) techniques have been integrated into computational materials science, enabling property prediction and accelerating materials analysis and design \cite{mozaffar2019deep,liu2021knowledge, guo2021artificial,buehler2022end,yang2022linking,buehler2022modeling, merchant2023scaling}. In early stages of such ML driven AI methods as applied to physics, prime applications have been as surrogate models or tools to serve specific inverse problems, like the design of new microstructures to meet certain material behaviors \cite{liu2022presto, buehler2023computational, buehler2023melm, lew2023single, lu2023modeling, ni2024forcegen, lu2024generative}. 

Materials design is inherently a multi-scale challenge, necessitating the integration of materials characteristics across different scales—from atomic interactions to macroscopic behaviors \cite{de2007multiscale, roters2010overview,  fish2021mesoscopic}. The vast array of data generated from these diverse scales—including deformation mechanisms, mechanical and thermal properties, processing-structure-property relationships, microstructural characteristics, and chemical compositions—exists in various formats such as text, images, and tabular data. Valuable insights are also embedded in resources like books, materials databases, patents, and technical reports \cite{buehler2008atomistic, jain2013commentary, de2015charting, anderson2017theory}. Moreover, designing materials with enhanced performance involves satisfying multiple criteria, a task complicated by the limitations of current modeling approaches. Physics-based models are precise but generally target specific properties and depend on computationally expensive input parameters derived from intensive simulations like DFT, and their integration requires reasoning over results, simulation strategies, and and understanding of relative weaknesses and strenghths of various tools. Machine learning and deep learning models complement these efforts in various ways: For instance, as surrogate models that bridge different scales enabling the exploration of massive design spaces \cite{lee2024data}, or by developing state-of-the-art ML interatomic potentials to achieve DFT accuracy with the speed of empirical potentials. \cite{behler2007generalized, kobayashi2017neural, mueller2020machine,  zuo2020performance, mishin2021machine}. 

However, a persistent frontier in materials science is the development of systems that cultivate comprehensive intelligence by automating complex materials modeling and design tasks, while leveraging a diverse array of knowledge, tools, and capabilities across different scales. These systems are crucial not only for generating novel insights into materials but also for dynamically integrating existing knowledge and developing new data. Importantly, these systems should ideally be able to iteratively refine their strategies, merging insights across disciplines to progressively evolve towards optimal solutions, along with an understanding, awareness and ability to probe specific details of the physics. Such a broad and integrative capability enables them to continuously enhance their approaches, adapting to new data (e.g. observations) and findings, thus improving the efficiency and effectiveness of material design and discovery processes. In this study, we propose a method to achieve such an integration of external knowledge, tools, logic, reasoning, and show how a physics-aware artificial intelligence can be effectively implemented through the deployment of a multi-model multi-agent AI system driven by Large Language Models (LLMs). 

Large language models \cite{wei2022emergent, chang2024survey} have demonstrated significant potential in various scientific and engineering domains \cite{boyko2023interdisciplinary, hu2023deep} such as materials \cite{jablonka202314, luu2024bioinspiredllm}, chemistry \cite{jablonka2024leveraging}, mechanics \cite{buehler2024mechgpt,buehler2023melm}, and proteins discovery \cite{madani2020progen}. Such models, built upon attention mechanism and transformer architectures \cite{vaswani2017attention} have shown proficiency in complex reasoning, strategic planning, coding, and workflow development, showing promising capabilities in materials analysis and prediction applications including hypothesis generation \cite{buehler2024accelerating}, critical knowledge retrieval via in-context learning, and multi-model reasoning \cite{buehler2024cephalo}. However, they face challenges in materials design due to limitations such as the inability to perform physics-based simulations, restricted access to external sources, and reliance on potentially outdated knowledge, which may not align with the rapidly advancing field of materials science.

\begin{figure*}[t]
\centering
\includegraphics[width=1\linewidth]{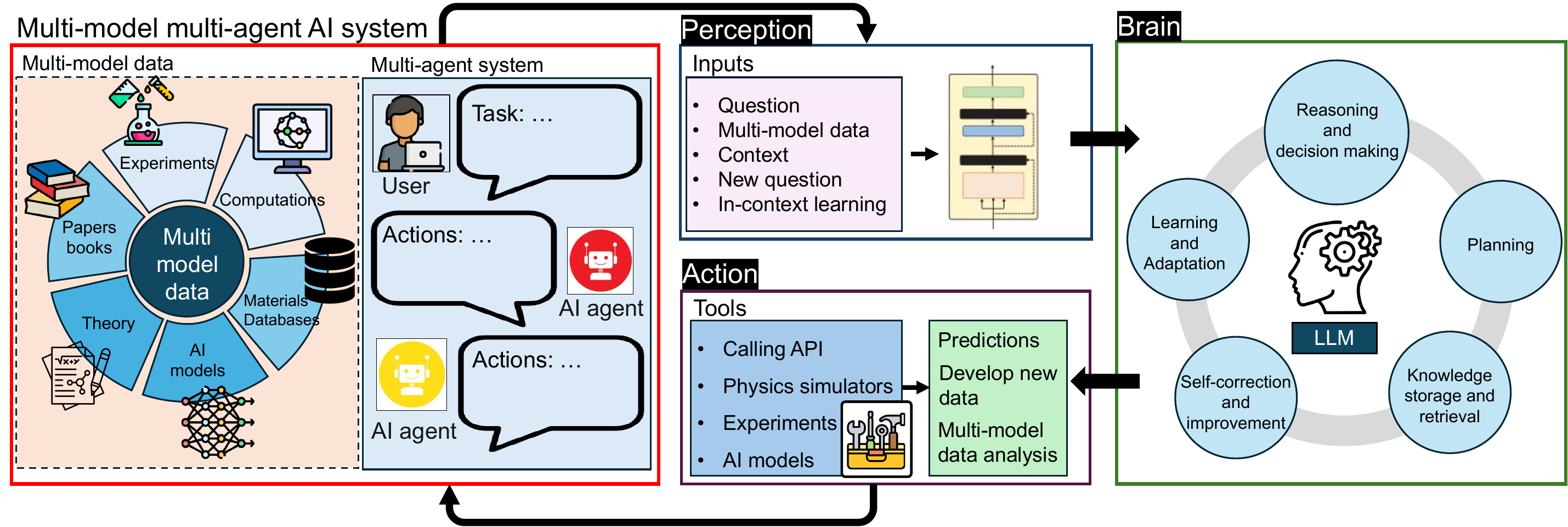}
\caption{\textbf{Multi-model multi-agent approach as a flexible modeling strategy for materials discovery, modeling, and prediction}. Multi-agent modeling can extend the power of large-language models by enabling the integration of multimodal data from diverse sources, including simulations, experiments, materials databases, and theoretical models.}
\label{fig:multi_agent_multidata}
\end{figure*}

To extend the utility of LLMs, we propose that multi-agent systems can serve as frameworks that transcend the traditional conversational functions of LLMs, with incipient applications across different domains \cite{Ni2023Agent, buehler2024generative, ghafarollahi2024protagents, m2024augmenting, stewart2024molecular}. 
These systems consist of three primary components \cite{Xi2023}: The brain, perception, and action, as illustrated in Figure \ref{fig:multi_agent_multidata}. The brain is the core of an AI agent model and is primarily composed of a frontier LLM, undertaking essential tasks of decision-making, reasoning, and planning. The perception module gathers and processes multi-model data, while the action module implements decisions based on the brain's guidance. The proposed workflow of our multi-model multi-agent system is as follows: first the perception module, perceives changes in the external environment and then converts multi-model information into an understandable representation for the agents. Next, the brain module, engages in processing activities such as planning, thinking, and decision-making. Finally, the action module, carries out the execution with the assistance of tools. By repeating the above process, the system of agents can collaborate, continuously get feedback and interact with the environment. 

The unique characteristics of multi-agent systems make them particularly suited to address the complexities of materials design, where traditional human-centric approaches may fail. These systems can integrate diverse data modalities and extract new knowledge from a broad range of external sources, including academic literature, online databases, and cutting-edge physics simulations, as depicted in Figure \ref{fig:multi_agent_multidata}. Enhanced by multi-modal LLMs, multi-agent systems are also capable of reasoning over images, facilitating the analysis of numerous experimental and computational results in visual formats. Such capabilities ensure the continuous evolution and improvement of material design and analysis, keeping pace with new scientific discoveries and methodological advancements. 

In this paper, we present a novel multi-agent approach designed to address the unique challenges associated with alloy modeling and design. We propose ``AtomAgents," a physics-aware multi-agent framework tailored to resolve complex issues in materials design that require detailed atomistic simulations. This framework utilizes a coordinated network of multi-agent systems to significantly enhance the efficiency and effectiveness of simulation processes in the development and analysis of crystalline materials at the atomic level. The primary contributions of our work are as follows:

\begin{itemize}
\item Integrating physics with generative AI: A deep capability to synergistically combine Large Language Models (LLMs) with detailed physics-based simulations, here demonstrated for the design of crystalline materials invoking the general-purpose LAMMPS MD code \cite{plimpton1995fast}.
\item Multi-Modal Data Integration: Our model has the capability to integrate multimodal data from various sources, enhancing its utility and adaptability in diverse research contexts.
\item Advanced Simulation Capabilities: Our model demonstrates exceptional performance in retrieving new physics through atomistic simulations, validated by several complex computational experiments.
\item Reduction in Human Intervention: AtomAgents significantly reduces the need for human intervention, displaying its capacity to autonomously design complex workflows, particularly useful for high-throughput simulations.
\item Accessibility for Non-Experts: Operating based on textual input, our model empowers non-expert researchers to effectively address challenges in the realm of crystalline materials design, making advanced simulations more accessible. 
\item Interpretability: The interactions between agents and tools is fully traceable for interpretation and analysis of intermediate results. This allows human researchers to understand potential issues and intervene, or redirect the process if necessary.

\end{itemize}

The structure of this paper is organized as follows. In Section \ref{sec:results}, we provide a detailed overview of the multi-agent system developed in our study. Subsequently, we present a series of computational experiments designed to demonstrate the efficacy of multi-agent collaboration in tackling complex tasks in alloy design and analysis involving atomistic simulations. Finally, a comprehensive discussion of the limitations and future perspectives of our approach is offered in Section \ref{sec:discussion}, featuring also a critical discussion of challenges and future research opportunities.

\section{Results and discussion}\label{sec:results}
\begin{figure*}[th!]
\centering
\includegraphics[width=1\linewidth]{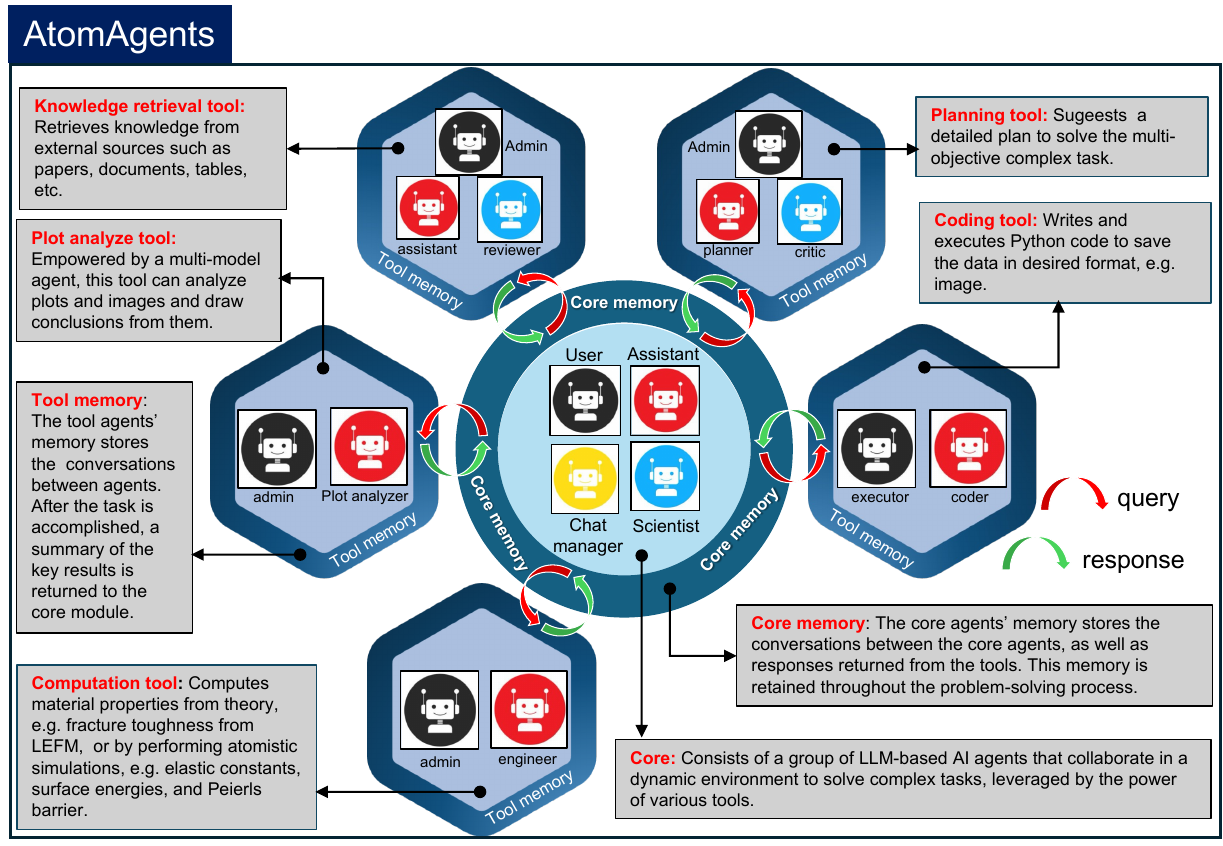}
    \caption{\textbf{AtomAgents, a physics-based generative multi-agent model for automating alloy discovery and analysis with atomistic simulations.} The structure of AtomAgents comprises a team of agents constructing the core who collaborate to solve complex alloy design tasks with the help of a set of tools for different purposes described in the image from knowledge retrieval to coding to image analysis. Each tool is composed of a set of AI agents that collaborate to solve the query received from the User and return the results to the core agents. Each individual AI agent in AtomtAgents is assigned a distinct profile that defines its role and may be powered by a general purpose large language model from the OpenAI GPT family. The entire process is automated, providing a robust framework for solving challenging tasks in alloy design and analysis with minimal or no human intervention.}
    \label{fig:AtomAgents}
\end{figure*}

The outline of our proposed multi-agent model is shown in Figure \ref{fig:AtomAgents}, illustrating the collaborative efforts of a team of agents to solve complex multi-objective problems in the context of alloy design and analysis requiring atomistic simulations. At the core of AtomAgents lies a group of agents—User, Engineer, Scientist, and Group Manager—who control the overall workflow of the problem-solving process by calling and executing relevant tools, providing appropriate inputs, and returning output results. Several tools are implemented in AtomAgents to facilitate the alloy design and analysis process, performing different tasks as shown in Figure \ref{fig:AtomAgents}. A fundamental tool is the planning tool, which consists of a group of agents: an Admin, a Planner, and a Critic, who are responsible for providing a detailed, well-structured plan to solve the multi-objective complex task. When a problem is posed, the planning tool is executed at the beginning, and a plan is created through collaboration between the Planner and the Critic, and returned to the core agents. Then, the core agents start executing the plan by calling the appropriate tools. These agents (except the User who poses the task) are powered by a state-of-the-art general-purpose large language model from the GPT family \cite{OpenAI2023} accessed via the OpenAI API \cite{OpenAI_API}. Moreover, each agent is characterized by a unique profile that describes its role in the system, as shown in Table \ref{tbl:agents_profile}. 

The full description of the tools and functions incorporated in AtomAgents is listed in Table S2 of the Supporting Information, showcasing a broad range of capabilities and functionalities. These include computations by atomistic simulations, knowledge retrieval from external sources, coding, plotting, and image analysis, collectively making AtomAgents a robust physics-aware LLM-based framework for solving intricate materials design and analysis tasks. As shown in Figure \ref{fig:AtomAgents}, each tool is equipped with a set of agents that autonomously collaborate to respond to the given query. Similar to the core agents, each of these AI agents is powered by a general-purpose LLM and is assigned a role described by its profile. Upon receiving a query from the user, the core agents dynamically and non-linearly call and execute the necessary tools until the solution is reached and the problem is resolved. The full profile of tool's agents is shown in Table S2 of the Supporting Information. Moreover, the computation tool is composed of a rich library of functions covering a wide range of atomistic simulations such as elastic constant and surface energy calculations, and performing complex nudged elastic band computations. The full list of computation functions along with their profile is listed in Table S3 in the Supporting Information.  

The following sections present a series of experiments to demonstrate how the multi-agent system addresses various tasks in the domain of alloy design, particularly through generating new physics \text{via} atomistic simulations, eliminating or substantially reducing the need for human intervention. All atomistic simulations are performed using LAMMPS \cite{plimpton1995fast} at zero temperature. Due to the weak performance of current state-of-the-art LLMs in constructing LAMMPS scripts, all the atomistic simulations are performed by human-generated LAMMPS scripts integrated into the model as python functions and executed \text{via} the computation tool. An example of such script is shown in Figure S1 in the Supplementary Information. 

The provided experiments highlight different strengths of the developed multi-agent system. Specifically, we show that the model can a) integrate materials properties from diverse sources,
b) tackle multi-modal problems involving image analysis, c) solve multi-scale problems connecting microscale feature to macroscopic properties, and generate and validate new hypotheses through atomistic simulations. These complex processes typically demand expertise from multiple domains, sophisticated reasoning abilities, and advanced scripting skills to execute atomistic simulations—tasks that have traditionally required significant manual effort. Our current approach showcases the capability of AI multi-agent modeling to not only seamlessly integrate different modalities but also to automate and optimize the entire workflow. By leveraging logic and reasoning capabilities of LLMs, the system can efficiently manage and solve intricate alloy design problems, significantly reducing the need for manual intervention. This innovation not only enhances the precision and effectiveness of alloy design processes but also accelerates discovery and development, paving the way for more advanced and efficient materials engineering.

\begin{table*}[th!]
 \centering
\caption{: The profiles of the agents implemented in the current study to solve multi-objective tasks in the context of alloy design and analysis.}
\begin{tabularx}{1.0\textwidth}{llX}
\toprule
\textbf{Agent \# } & \textbf{Agent role} & \textbf{Agent profile} \tabularnewline
\midrule
1 (None) & User &  User. You pose the task. \tabularnewline
\midrule
2 (GPT-4o)& Assistant & Assistant. You execute a plan developed by Planning tool.
\tabularnewline
\midrule
3 (GPT-4-turbo)& Scientist & Scientist. You are a leading material scientist. You propose innovative hypotheses in the realm of materials science.  
\tabularnewline
\midrule
4 (GPT-4o)& Group chat manager &  You dynamically select a speaker.
\tabularnewline
\midrule
5 (GPT-4-turbo)& Planner &  Planner. Suggest a plan to solve a problem. Do not execute the plan. The plan should include the functions and required input parameters. The plan should be approved by the critic. 
\tabularnewline
\midrule
5 (GPT-4-turbo)& Critic &  Critic. Double-check the plan from planner for completeness and correctness. Ensure the plan includes all the functions with correct input parameters.
\tabularnewline
\bottomrule
\end{tabularx}
\label{tbl:agents_profile}
\end{table*}

\subsection{Experiment I - Material Properties Calculation and Knowledge Retrieval; multi-model integration problem}\label{sec:exp_1}
Computing material properties such as lattice and elastic constants is crucial in atomistic simulations. These properties characterize materials and are particularly important for theoretical modeling. For instance, when computing the critical fracture toughness based on Griffith's theory, both the elastic constants and surface energies must be determined. However, computing various material properties often requires setting up different structures, writing and adjusting multiple LAMMPS scripts, and running numerous simulations, which becomes cumbersome when studying a broad range of materials, especially in alloy design. 

\begin{figure}[tbht]
\centering
        \includegraphics[width=.75\textwidth]{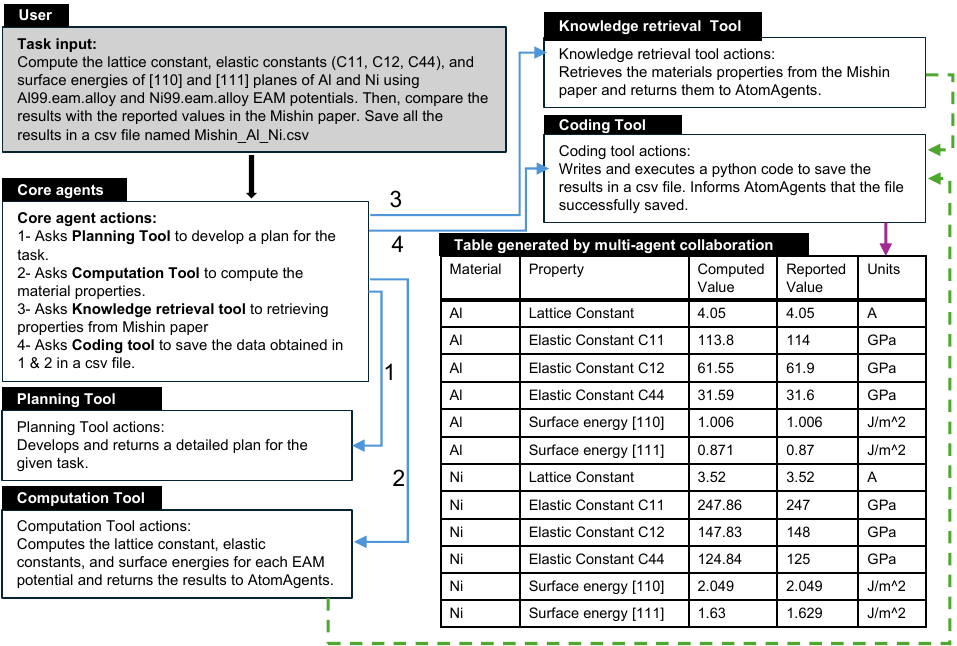}
    \caption{Overview of the multi-agent collaboration to solve the complex task posed in Experiment I. After receiving the task from the user, the core agents call the "planning" tool to create a plan for the task. Then the core agents start executing the plan by using ``computation" tool to compute the material properties and ``knowledge retrieval" tool to retrieve the material properties from a set of scientific papers or other documents. Finally, all the data are collected and sent to ``Coding" tool to save them in a comma-separated values (CSV) file.}
    \label{fig:exp_1_work}
\end{figure}

In this section, we demonstrate the capability of multi-agent modeling to perform complex tasks without the need for extensive coding knowledge or expertise in LAMMPS. Furthermore, we illustrate how multi-agents can automate the extraction of valuable knowledge from the literature, such as material properties computed by specific interatomic potentials from corresponding papers.  
The entire problem-solving process is managed by AI agents, encompassing plan definition, simulation execution, knowledge retrieval, and the storage of results. The workflow illustrating multi-agent collaboration for this experiment is shown in Figure \ref{fig:exp_1_work}.

Initially, the core agents activate the "planning" tool responsible for developing a plan to address the query posed by the user. The "planner" agent within this tool breaks down the complex task into simpler sub-tasks by proposing a step-by-step plan. This plan is then evaluated and approved by the accompanying "critic" agent, and the approved plan is returned to the AtomAgents for execution. The plan involves using a computation tool to derive material properties from atomistic simulations, a knowledge retrieval tool to extract these properties from papers, and coding tools to write a Python script for saving the results.

The ``Assistant" agents in AtomAgents demonstrate excellent performance in following the developed plan, calling, and executing the relevant functions. They are particularly adept at providing the correct query to the team of agents embedded in different tools, as these agents do not have access to external information. Once the results from the computation and knowledge retrieval tools are returned, they are collected and formatted as a Python dictionary. This dictionary is then utilized by the ``coder" agent to save the results in a CSV file, as illustrated in Figure \ref{fig:exp_1_work} in tabular format.

This experiment showcases the efficacy of the multi-agent system in solving complex tasks that involve conducting atomistic simulations. Moreover, the agreement between the computed and reported values confirms the accuracy of the computations, paving the way for more challenging experiments.

\subsection{Experiment II-Analyzing screw dislocation core structure; multi-model analysis problem}
The previous experiment focused on computing basic material properties through atomistic simulations conducted on pristine materials. In this experiment, we extend the scope to include simulations on defected structures, specifically modeling screw dislocations in BCC materials. Screw dislocations are line defects that significantly influence the plasticity of BCC materials and have been extensively studied through atomistic simulations. A critical aspect of these studies is the dislocation core structure and its interaction with nearby solutes. Many empirical potentials suggest that the screw dislocation core in pure BCC metals is polarized, whereas ab-initio simulations based on Density Functional Theory (DFT) indicate an unpolarized, compact core structure. The nature of the core structure profoundly affects the movement of long screw dislocations and is crucial for accurately investigating dislocation motion mechanisms, particularly double-kink nucleation. To evaluate the performance of these potentials in predicting the core structure accurately, we analyze differential displacement maps. These maps are generated by subtracting the atomic positions in dislocated structures from those in pristine conditions. Consequently, this experiment employs a specialized multi-model agent equipped with image reasoning capabilities, enhancing the model's performance ability to accurately characterize and understand these complex defects.

\begin{figure}[h]
        \centering
        \includegraphics[width=.75\textwidth]{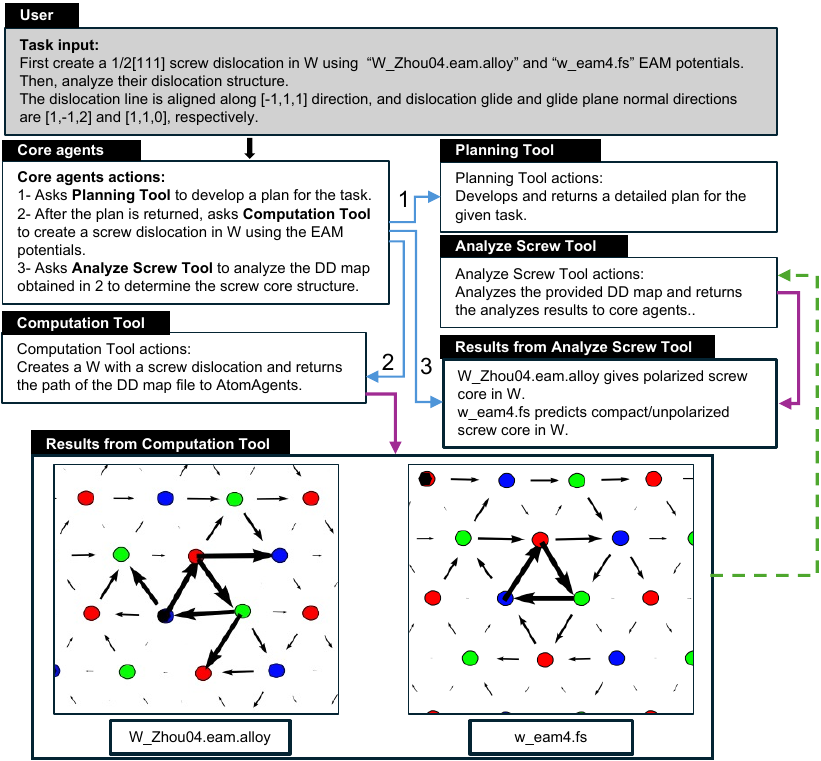}
    \caption{Overview of the multi-agent collaboration to solve the complex task posed in Experiment II. After receiving the task from the user, the core agents call the "planning" tool to create a plan for the task. Then the core agents start executing the plan by using ``computation" tool to generate the DD maps and subsequently ``Analyze Screw" tool to determine the screw core structure. The DD maps returned by the ``computation" tool and the summary of the image analysis made by the "Analyze Screw" tool are included in the image.}
    \label{fig:exp_2_work}
\end{figure}

In this segment, we address a complex multi-objective problem where a team of agents is tasked with determining the screw dislocation core structure in BCC tungsten using two commonly employed EAM potentials: Zhou-Johnson \cite{zhou2004misfit} and Marinica (EAM4) \cite{marinica2013interatomic}. Figure \ref{fig:exp_2_work} illustrates the workflow executed by the agents to conduct this experiment. The process begins with a query from the user and proceeds through meticulous planning, execution of atomistic simulations, and culminates in image analysis to examine the generated differential displacement (DD) maps. The plan ensures the correct tools are executed with the appropriate input parameters. Furthermore, the computation tool performs all necessary tasks to create the differential displacement maps without errors. This includes creating the potential file describing pair\_style and pair\_coeff parameters, computing the lattice constant required to create the BCC structure, creating and relaxing a screw dislocation within tungsten, and constructing the differential displacement maps as shown in Figure \ref{fig:exp_2_work}.

The last and crucial step in this experiment is the determining of the core structure of the screw dislocations from the generated DD maps. This is done by our multi-model agent capable of reasoning over images. However, our initial assessments indicated the poor performance of the multi-model agent equipped with GPT-4o (or GPT-4v) in distinguishing between the polarized and unpolarized core structures. To circumvent this, we leverage in-context learning to empower the multi-model agent to correctly identify screw core structure by providing it an image including polarized and unpolarized cores with corresponding labels as illustrated in Figure S2 in the Supporting Information. 

The results from the "Analyze Screw Core" tool indicate that the multi-model agent has successfully identified the screw dislocation core structure in W generated by the two eam potentials. After the successful completion of all the computation and analysis tasks, a summary of the final results is returned by the "Assistant" agent:
\begin{Box1}[colbacktitle={black!20!white}, colback={black!10!white}]{Assistant}
\footnotesize{\texttt{The analysis of the screw dislocation core structures for the 1/2[111] screw dislocation in Tungsten (W) using two different potentials has been completed:
\\
1. **Using W\_Zhou04.eam.alloy Potential:**\\
\noindent\hspace*{0.45cm}- The screw dislocation core structure is identified as a **polarized/non-compact core**.\\
\noindent\hspace*{0.45cm}- The displacement vectors around the core form a more complex, non-triangular pattern, indicating a non-compact core structure.
\\
2. **Using w\_eam4.fs Potential:**\\
\noindent\hspace*{0.45cm}- The screw dislocation core structure is identified as a **compact/unpolarized core**.\\
\noindent\hspace*{0.45cm}- The displacement vectors are directed towards the center of a triangular arrangement, indicating a compact core structure.}}
\end{Box1}  
The collaborative efforts of the agents in this experiment, leveraging computational techniques and multi-model reasoning capabilities, ensure a thorough and precise evaluation of the dislocation core structures as predicted by the selected potentials. 

\subsection{Experiment III - Multi-Scale Mechanics Problem: Fracture Toughness in Alloy Systems}
To further explore the capabilities of our multi-agent model in solving complex simulation-based problems, this experiment is devoted to addressing a multi-scale mechanics problem that combines simulations with theories. Specifically, this task involves computing the critical fracture toughness, a macro-scale material property. Based on linear elastic fracture mechanics (refer to  \ref{sec:fracture_toughness} for more details), computing the critical fracture toughness necessitates calculating the elastic constants and surface energies of the material. Moreover, this task extends beyond computations in pure materials, as conducted in previous experiments, by incorporating simulations on alloy systems.

\begin{figure}[h]
        \centering
        \includegraphics[width=.75\textwidth]{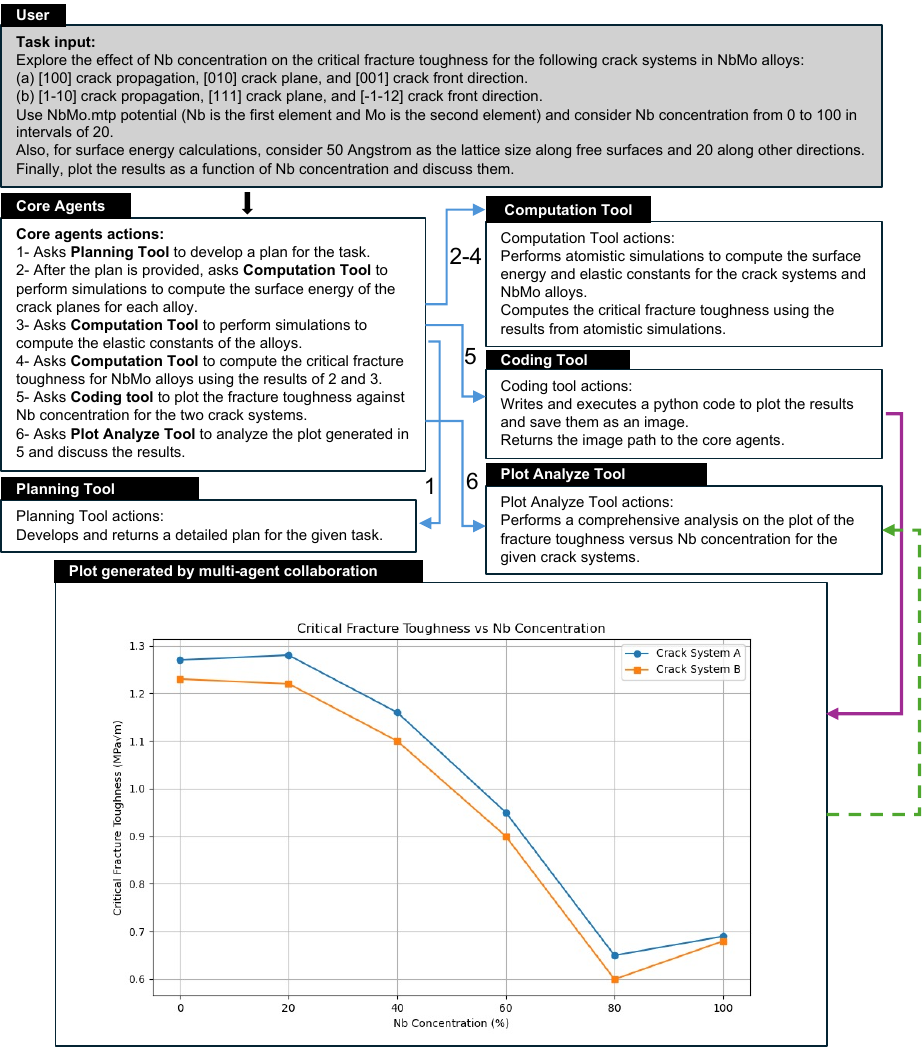}
    \caption{Overview of the multi-agent collaboration to solve the complex task posed in Experiment III. After receiving the task from the user, the core agents call the ``planning" tool to create a plan for the task. Then the core agents start executing the plan by using "computation" tool to compute the elastic constants and surface energies, and eventually the fracture toughness. Subsequently ``Coding" tool is used to plot the results which are then analyzed by the ``Plot analyze" tool via a multi-model agent.}
    \label{fig:exp_3_work}
\end{figure}

The overall workflow employed by our multi-agent system is illustrated in Figure \ref{fig:exp_3_work}, where the user poses a highly complex multi-objective task to explore the effect of niobium (Nb) concentration on the fracture toughness in NbMo alloys for two crack systems. An important feature of this experiment is the use of a moment tensor potential \cite{shapeev2016moment, yin2021atomistic}, a machine-learning-based potential, to describe the interatomic interactions, showcasing the flexibility of the multi-agent system. Furthermore, the task includes plotting the results and discussing them, which is performed by the multi-model agent expert in analyzing images and plots. Another critical aspect of this problem is the computation of material properties (elastic constants and surface energies) for alloy systems. Due to fluctuations in random solute environments, multiple configurations may need to be sampled to obtain an average material property, increasing the computational cost. Here, we have assigned five as the default value for the number of samples, but the user can opt for higher values if needed (e.g., user assigns 20 samples for the NEB simulations in Experiment IV).

At the initial step, the core agents engage the ``planning" tool to provide a detailed implementation plan for the given task. The ``planner" agent develops a comprehensive plan, which is then validated for accuracy by the ``critic" agent. The plan (shown in Figure S3 in the supporting information), demonstrates the planner's effectiveness in identifying the key steps necessary for accomplishing the tasks posed by the user. Specifically, the planner correctly identifies that calculating the critical fracture toughness requires computing the surface energy and elastic constants. Moreover, it accurately recognizes that the surface energy should be computed along the crack planes. Saving the final results and analyzing them is another crucial step adeptly captured in the plan.

Upon receiving the plan, the "Assistant" agent in the core module initiates the implementation of the plan step-by-step by activating the tools as outlined. Despite the complexity and length of the tasks, all tools are implemented and the necessary results are gathered autonomously, without human intervention. After the material properties are collected, the fracture toughness is computed for the specified crack systems and alloy compositions. These results are then used as input for the plotting tool.

\begin{figure}[ht!]
    \centering
\begin{Box1}[colbacktitle={white!20!white}, colback={black!10!white}]{Plot analyze tool (GPT-4o)}
\footnotesize{\texttt{The plot shows the variation of critical fracture toughness as a function of Nb concentration for two crack systems (A and B) in NbMo alloys. Here are some conclusions that can be drawn from the results:}}
\begin{wrapfigure}{r}{0.6\textwidth}
  \begin{center}
 \includegraphics[width=0.58\textwidth]{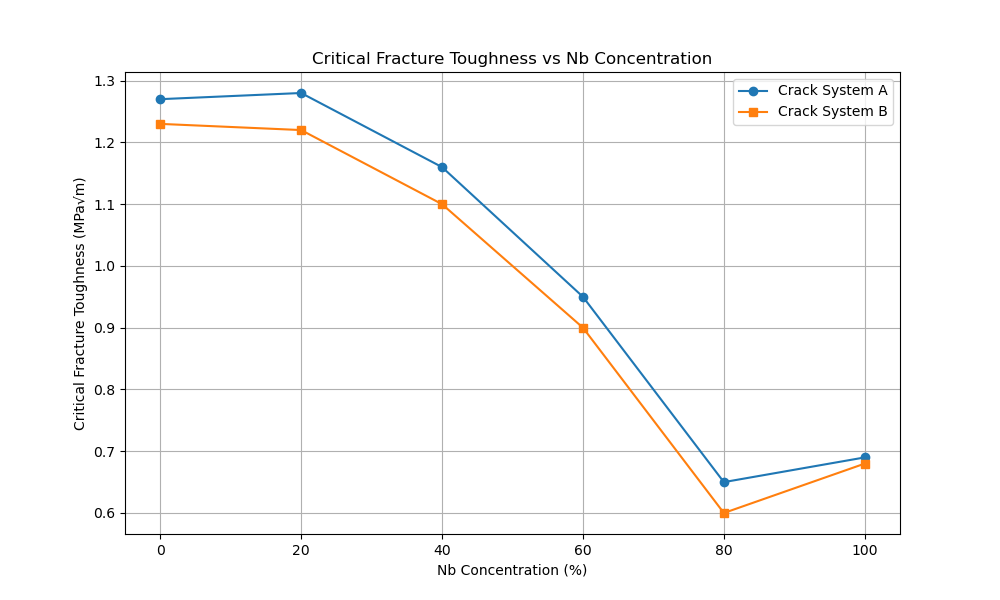}
% \caption{Critical fracture toughness versus Nb concentration computed by the %autonemous multi-agent collaboration for the experiment III.}
   \end{center}
 \end{wrapfigure}
 \footnotesize{\texttt{
1. **General Trend**:\\
\noindent\hspace*{0.45cm}- For both crack systems, the critical fracture toughness generally decreases as the Nb concentration increases from 0\% to 80\%.\\
\noindent\hspace*{0.45cm}- There is a slight increase in fracture toughness at 100\% Nb concentration for both crack systems.\\
2. **Crack System A**:\\
\noindent\hspace*{0.45cm}- At 0\% Nb concentration, the critical fracture toughness is approximately 1.25 MPa$\sqrt{m}$.\\
\noindent\hspace*{0.45cm}- The fracture toughness remains relatively stable up to 20\% Nb concentration, with a slight increase.\\
\noindent\hspace*{0.45cm}- Beyond 20\% Nb concentration, the fracture toughness decreases steadily, reaching a minimum at 80\% Nb concentration (around 0.65 MPa$\sqrt{m}$).\\
\noindent\hspace*{0.45cm}- At 100\% Nb concentration, there is a slight increase in fracture toughness to approximately 0.75 MPa$\sqrt{m}$.\\
3. **Crack System B**:\\
\noindent\hspace*{0.45cm}- At 0\% Nb concentration, the critical fracture toughness is approximately 1.2 MPa$\sqrt{m}$.\\
\noindent\hspace*{0.45cm}- The fracture toughness decreases steadily as Nb concentration increases, reaching a minimum at 80\% Nb concentration (around 0.6 MPa$\sqrt{m}$).\\
\noindent\hspace*{0.45cm}- At 100\% Nb concentration, there is a slight increase in fracture toughness to approximately 0.7 MPa$\sqrt{m}$.\\
\\
4. **Comparison Between Crack Systems**:\\
\noindent\hspace*{0.45cm}- Crack System A generally exhibits higher critical fracture toughness than Crack System B across all Nb concentrations.\\
\noindent\hspace*{0.45cm}- The difference in fracture toughness between the two crack systems is more pronounced at lower Nb concentrations (0\% to 40\%) and becomes less significant at higher Nb concentrations (60\% to 100\%).\\
\\
5. **Implications**:\\
\noindent\hspace*{0.45cm}- The decrease in fracture toughness with increasing Nb concentration suggests that higher Nb content may lead to a more brittle material.\\
\noindent\hspace*{0.45cm}- The slight increase in fracture toughness at 100\% Nb concentration for both crack systems could indicate a change in the fracture mechanism or material properties at this composition.\\
\\
These observations can help in understanding the mechanical behavior of NbMo alloys and in optimizing the Nb concentration for desired fracture toughness properties.}}
\end{Box1}
    \caption{The analyzes returned by the "plot analyzer" agent for the plot of critical fracture toughness versus Nb concentration created by the multi-agent collaboration.}
    \label{fig:anlyses_exp3}
\end{figure}

The plot of the results, generated by the multi-agent collaboration, is displayed in Figure \ref{fig:exp_3_work}. Subsequently, the plot is analyzed, and conclusions are drawn by the multi-model agent within the analysis tool, as shown in Figure \ref{fig:anlyses_exp3}. The multi-model agent, empowered by GPT-4, adeptly captures all details of the plot both quantitatively and qualitatively. This includes identifying the maximum and minimum values of the fracture toughness for the crack systems and providing a detailed discussion on the variation of fracture toughness relative to the Nb concentration.

The results from this experiment showcase the competence of the multi-model agent in providing a robust framework for automating the design of alloys. This integration effectively combines various modalities such as theory, simulations, coding, and image analysis, illustrating the comprehensive capabilities of the multi-agent system.

\subsection{Experiment IV-Hypothesis generation and validation}
The previous experiments demonstrated the exceptional performance of our multi-agent model in solving complex computational problems requiring atomistic simulations. In this experiment, we explore another intriguing aspect of the multi-agent strategy: hypothesis generation and validation. Due to their high reasoning capabilities, general-purpose large language models like the GPT family have the potential to generate innovative research ideas across fields, including materials science. Combining this with the computational prowess of the multi-agent strategy offers a robust framework where LLM-based research ideas and hypotheses can be validated and refined, providing new opportunities for scientific discovery. The objective of this experiment is to assess the performance of our multi-agent model in hypothesis generation and validation.

It is well-known that plastic deformation in BCC metals is dominated by <111> screw dislocation motion. An important physical property of BCC materials is the Peierls potential, which is the energy change of the straight dislocation line as it moves from one Peierls valley (minimum energy configuration) to the next. The maximum energy along this profile is the Peierls barrier, which profoundly affects screw dislocation motion and, consequently, the mechanical properties of BCC materials. Therefore, accurately estimating the Peierls potential is essential for predicting and understanding the behavior of BCC materials. The Peierls barrier can be computed using transition state calculations such as the nudged elastic band (NEB) method. While the Peierls potential in pure BCC metals is constant and has been extensively studied \cite{proville2012quantum, groger2013stress, ventelon2013ab, dezerald2014ab, rodney2017ab, maresca2018screw}, the energy landscape in alloys is more complex, leading to a wide distribution of energy barriers. Consequently, multiple NEB simulations are required to accurately estimate the mean barrier. However, this process becomes computationally expensive when screening the entire design space, for instance, to search for compositions with high energy barriers. Therefore, finding a correlation between the Peierls barriers and more easily computed properties could significantly save time by narrowing down the composition space towards designs with higher barriers and thus enhanced mechanical performance.

\begin{figure}[ht!]
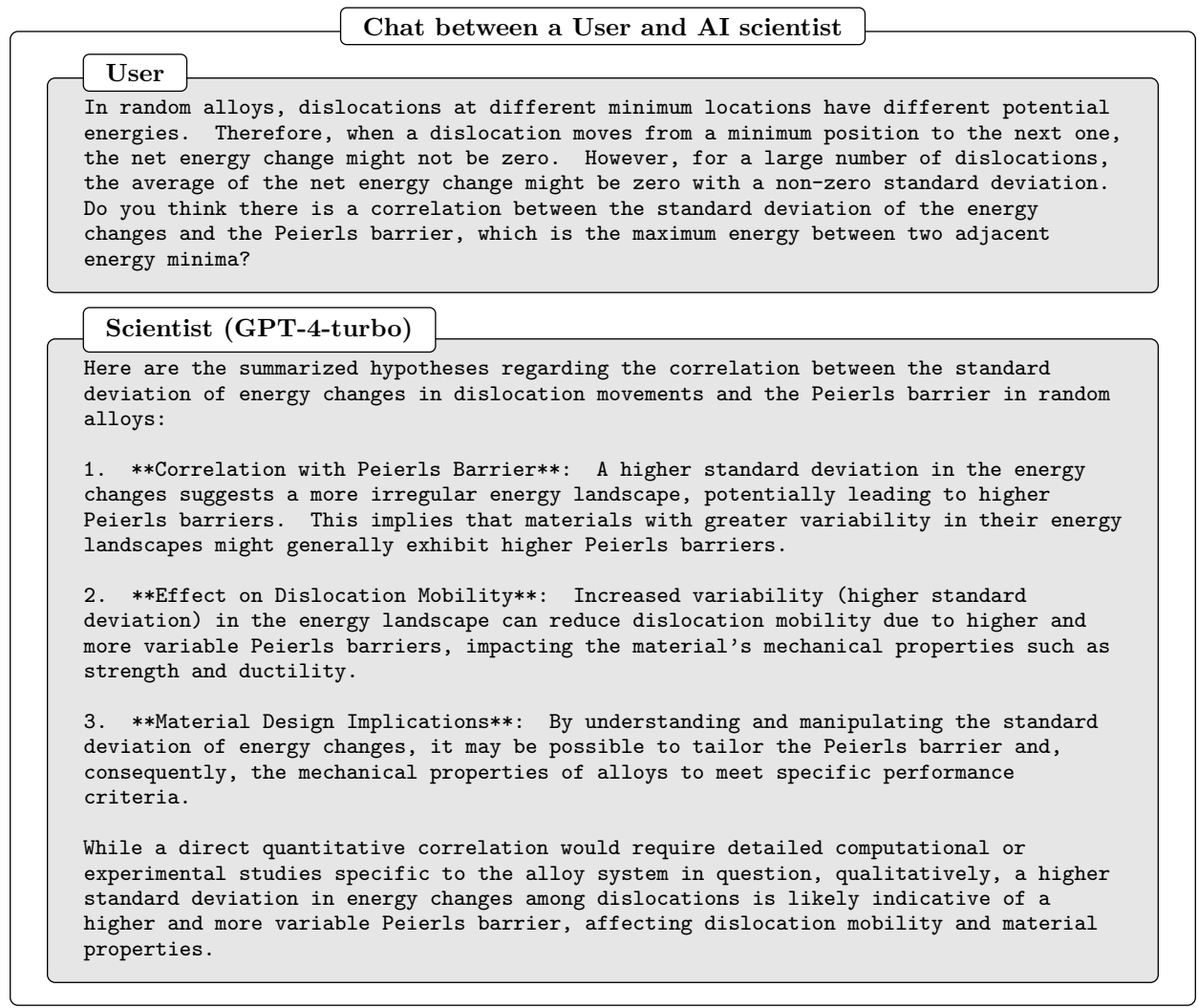

    \centering
    \begin{Box2}[colbacktitle={white!20!white}, colback={white!10!white}]{Chat between a User and AI scientist}
\begin{Box1}[colbacktitle={white!20!white}, colback={black!10!white}]{User}
\footnotesize{\texttt{In random alloys, dislocations at different minimum locations have different potential energies. Therefore, when a dislocation moves from a minimum position to the next one, the net energy change might not be zero. However, for a large number of dislocations, the average of the net energy change might be zero with a non-zero standard deviation. Do you think there is a correlation between the standard deviation of the energy changes and the Peierls barrier, which is the maximum energy between two adjacent energy minima?}}
\end{Box1}
\begin{Box1}[colbacktitle={white!20!white}, colback={black!10!white}]{Scientist (GPT-4-turbo)}
\footnotesize{\texttt{Here are the summarized hypotheses regarding the correlation between the standard deviation of energy changes in dislocation movements and the Peierls barrier in random alloys:\\
\\
1. **Correlation with Peierls Barrier**: A higher standard deviation in the energy changes suggests a more irregular energy landscape, potentially leading to higher Peierls barriers. This implies that materials with greater variability in their energy landscapes might generally exhibit higher Peierls barriers.\\
\\
2. **Effect on Dislocation Mobility**: Increased variability (higher standard deviation) in the energy landscape can reduce dislocation mobility due to higher and more variable Peierls barriers, impacting the material's mechanical properties such as strength and ductility.\\
\\
3. **Material Design Implications**: By understanding and manipulating the standard deviation of energy changes, it may be possible to tailor the Peierls barrier and, consequently, the mechanical properties of alloys to meet specific performance criteria.\\
\\
While a direct quantitative correlation would require detailed computational or experimental studies specific to the alloy system in question, qualitatively, a higher standard deviation in energy changes among dislocations is likely indicative of a higher and more variable Peierls barrier, affecting dislocation mobility and material properties.}}
\end{Box1}
\end{Box2}
    \caption{The hypotheses generated by the "scientist" agent empowered by GPT-4-turbo in response to the user's query.}
    \label{fig:scientist_hypotheses}
\end{figure}

To facilitate hypothesis generation, a ``Scientist" agent is integrated into our model, profiled to act as a leading materials scientist. In the current framework, the user can directly collaborate with the Scientist to generate innovative hypotheses. However, it is possible to integrate other mechanisms for hypothesis generation, such as two AI agents that adversarially collaborate to push each other towards more innovative and robust research ideas, or extracting and generating novel research ideas from powerful knowledge graphs \cite{buehler2024generative, buehler2024accelerating, buehler2024x}. Figure \ref{fig:scientist_hypotheses} shows a summary of the hypotheses generated by the "Scientist" agent in response to the user's query. It can be observed that, the scientist proposes a positive correlation between the Peierls barrier and the standard deviation of the energy change. Th proposed correlation by the ``Scientist" is particularly encouraging as it reduces the computational burden of performing costly NEB simulations to compute the Peierls barrier directly.

\begin{figure}[h]
        \centering
        \includegraphics[width=.75\textwidth]{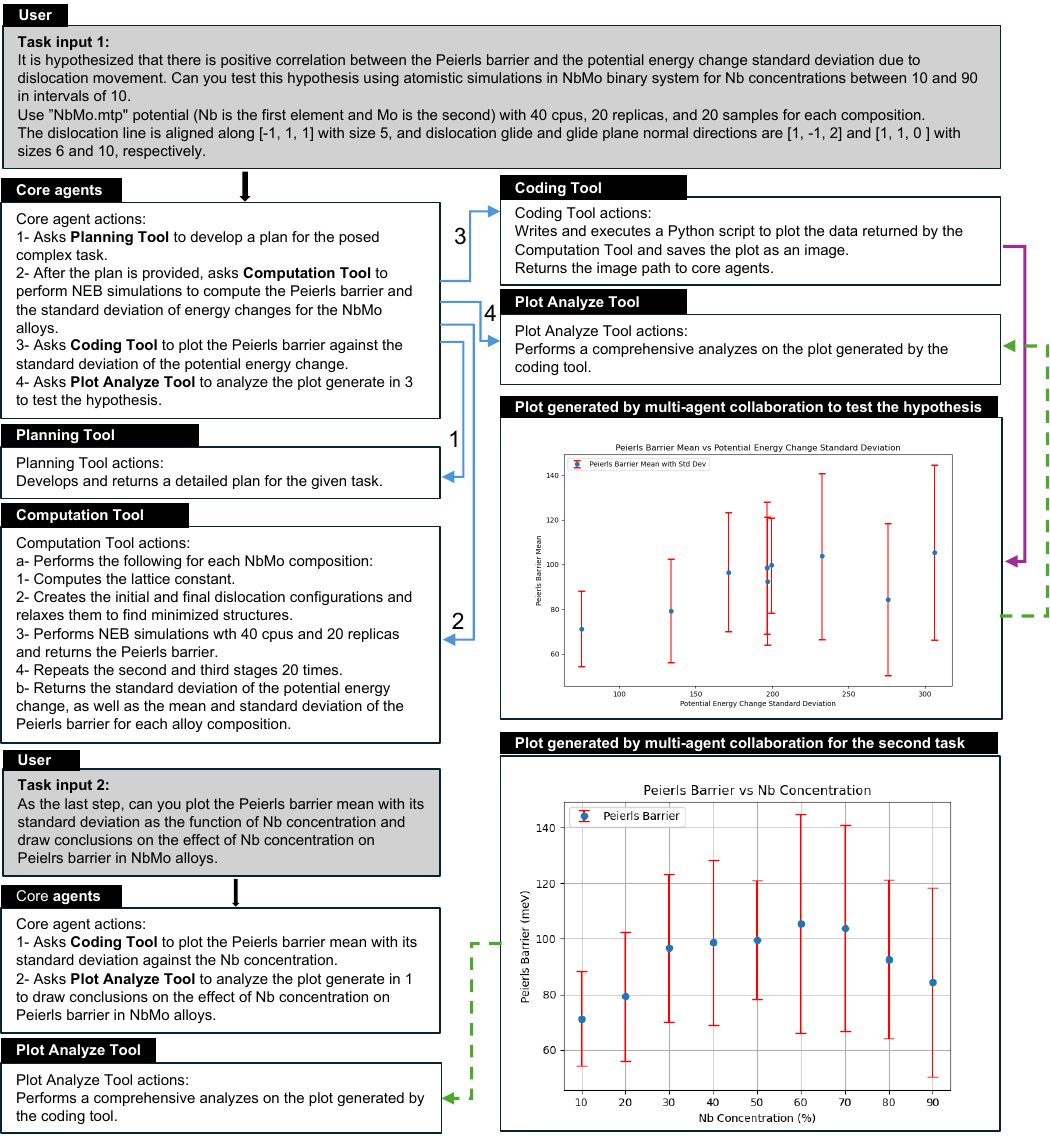}
    \caption{Overview of the multi-agent collaboration to solve the complex task posed in Experiment IV and then a follow-up task. After receiving the task from the user, the core agents call the "planning" tool to create a plan for the task. Then the core agents start executing the plan by using "computation" tool to compute the Peierls barrier and the standard deviation of the potential energy changes for a set of alloys. Subsequently "Coding" tool is used to plot the results which are then analyzed by the "Plot analyze" tool via a multi-model agent. In the second task, the results of the first task are plotted via the coding agent to show the variation of the Peierls barrier against Nb concentration. The final analysis from the multi-agent model finalizes the task.}
    \label{fig:exp_4_work}
\end{figure}

We now utilize our multi-agent system to test the hypotheses generated in the previous experiments using atomistic simulations. As in the earlier experiment, we use NbMo as our model system, described by the moment tensor potential, though it is feasible to apply other alloy systems and interatomic potentials. The overall workflow of the problem-solving process is depicted in Figure \ref{fig:exp_4_work}, where the user assigns a computational task pertaining to the aforementioned hypotheses, followed by a detailed plan from the "Planner." Subsequent computations are performed by the "Computation Tool" to compute the energy change and Peierls barrier from NEB simulations. The details of these NEB simulations can be found in \ref{sec:methods}.

\begin{figure}[h!]
    \centering
\begin{Box1}[colbacktitle={white!20!white}, colback={black!10!white}]{Plot analyze tool (GPT-4o)}
\footnotesize{\texttt{To draw conclusions from the data presented in the plot, we need to analyze the relationship between the Peierls barrier mean and the potential energy change standard deviation. Here are some observations and conclusions:}}
\begin{wrapfigure}{r}{0.6\textwidth}
  \begin{center}
 \includegraphics[width=0.58\textwidth]{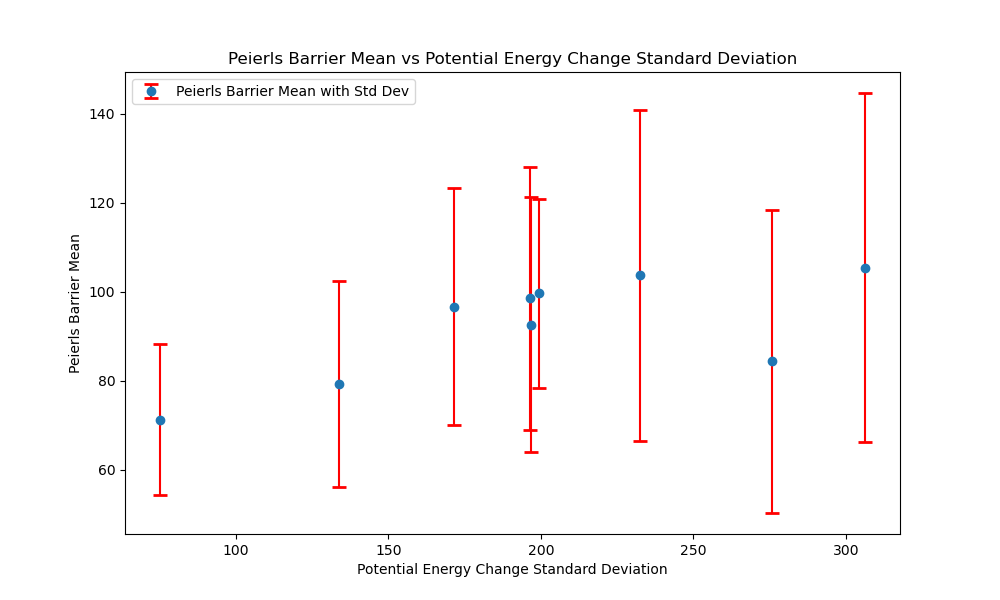}
% \caption{Critical fracture toughness versus Nb concentration computed by the %autonemous multi-agent collaboration for the experiment III.}
   \end{center}
 \end{wrapfigure}
\footnotesize{\texttt{1. **Positive Correlation**:\\ 
   - The plot shows that as the potential energy change standard deviation increases, the Peierls barrier mean also tends to increase. This suggests a positive correlation between the two variables.\\
2. **Error Bars**:\\
   - The error bars represent the standard deviation of the Peierls barrier mean. The error bars are relatively large, indicating significant variability in the Peierls barrier mean for each potential energy change standard deviation.\\
3. **Trend Analysis**:\\
   - Despite the variability, there is a noticeable upward trend in the Peierls barrier mean with increasing potential energy change standard deviation. This trend supports the conclusion of a positive correlation.\\
4. **Data Points**:\\
   - The data points are scattered but generally follow an upward trajectory. This further reinforces the positive correlation between the Peierls barrier mean and the potential energy change standard deviation.\\
In summary, the plot indicates a positive correlation between the Peierls barrier mean and the potential energy change standard deviation for different compositions of the NbMo binary system. However, the significant variability (as shown by the error bars) suggests that while the trend is positive, there is considerable spread in the data.}}
\end{Box1}
    \caption{The results of the analyze tool from the analyzes of the plot generated by the multi-agent collaboration in experiment IV.}
    \label{fig:exp_4_analyze_1}
\end{figure}

After the completion of the NEB simulations, the results are collected and sent to the "Coding Tool" for plotting. Figure \ref{fig:exp_4_work} presents the plot of the Peierls barrier mean and standard deviation against the potential energy change standard deviation. The analysis of this plot is shown in Figure \ref{fig:exp_4_analyze_1}. The comprehensive analysis reveals the proficient performance of the multi-model agent, empowered by GPT-4o, in analyzing the plot. Following a thorough review, the agent concludes that there is a positive correlation between the mean Peierls barrier and the standard deviation of the potential energy change, hence verifying the hypothesis developed by the "Scientist". This finding is significant as it points toward a method for accelerating the design process of materials with enhanced mechanical performance by focusing on specific energy characteristics that are indicative of higher barriers.

AtomAgents facilitates human-AI collaboration at various stages of the problem-solving process. For example, after the multi-agent system successfully completed the complex computations and analyses required to test the hypotheses, we posed another task to explore the effect of niobium (Nb) concentration on the Peierls barrier in NbMo alloys. Since the necessary data had already been collected during the main task, no further simulations were required. Figure \ref{fig:exp_4_work} displays the follow-up task posed by the user to the multi-agent system, along with the plot generated by the ``Coding Tool."
This plot is subsequently evaluated by the ``Analyze Plot" tool to draw conclusions about the impact of the Nb concentration on the Peierls barrier in NbMo alloys. The results of this analysis are depicted in Figure S4 in the supporting material, demonstrating the proficiency of the multi-model agent, powered by GPT-4o, in analyzing the variations of the Peierls barrier with respect to the Nb concentration. This capability highlights the system's effectiveness in leveraging existing data to generate new insights without the need for additional computational resources, thus streamlining the research process and enhancing the efficiency of materials design evaluations.

\section{Summary and future perspective}\label{sec:discussion}

We constructed a physics-aware artificial intelligence (AI) model that combines multimodal capabilities in complex reasoning, rational thinking, and strategic planning of large language models (LLMs) with abilities to write and execute code, conduct atomistic-level molecular simulations to solicit new real-world physics data, and conduct visual analysis of graphed data, molecular mechanisms and motions. Through a multi-agent strategy in which these capabilities are integrated, we construct a physics-aware intelligent system capable of solving complex analysis and design tasks, applied here to alloy design and discovery.
We showed that such an integration is possible to model complex compositional problems in alloy modeling and design.  Our model, AtomAgents, combines a comprehensive suite of tools with various capabilities, including physics-based atomistic simulators, effectively coupling the formidable multimodal reasoning capabilities of LLMs with the essential computational power required for addressing sophisticated materials design problems.

A key feature of the proposed multi-model system is the potential of integration of different modalities from different domains into the materials design process, offering a paradigms shift in solving challening multi-scale materials problems that traditionally demands considerable human expertise and manual effort. The comprehensive data integration allows AtomAgents to continuously update and improve its predictive capabilities, ensuring that the latest findings and trends in materials science are incorporated into its models. The systematic assimilation of such diverse and dynamic data sources not only enhances the accuracy of simulations and predictions but also accelerates the materials discovery process, paving the way for groundbreaking advancements in the field

While the research reported here primarily focuses on designing binary alloys and metallic materials, the AtomAgents framework inherently offers the flexibility to explore more complex systems, such as high-entropy alloys~\cite{miracle2017critical, george2019high, george2019high}. These systems open a vast compositional design space, providing unprecedented opportunities to engineer materials with novel properties like enhanced strength and fracture toughness \cite{ghafarollahi2019solute, rao2022theory}. However, navigating this extensive space presents significant challenges, requiring innovative and precise methodologies to manage its complexities effectively. AtomAgents, with its multi-agent architecture, is ideally equipped for these tasks, utilizing collective intelligence to efficiently explore and exploit this expansive design space. Furthermore, our framework is not confined to just metallic materials; owing to the versatility of LAMMPS, which serves as our physics engine, AtomAgents can also model a wide range of other materials systems—including biomaterials, polymers, ceramics, and liquids, that can be readily simulated within the LAMMPS environment.

Further, the potential of integrating advanced deep learning models and generative tools within AtomAgents offers a promising avenue to enhance its capabilities. These AI-driven models can predict material properties across diverse alloy systems, potentially reducing reliance on extensive atomistic simulations \cite{fan2022predicting,wang2024neural}. Such integration could significantly streamline the process, reducing both time and computational overhead, especially when dealing with complex multi-component alloys. This strategic enhancement would not only optimize the efficiency of material design but also expand the boundaries of what can be achieved in materials science, propelling AtomAgents to the forefront of innovative materials discovery.

Working as the integrative agent in the multi-agent system, Large Language Models (LLMs) play a crucial role in orchestrating the interactions and operations within these systems. They facilitate critical steps such as planning, reasoning, and critical thinking, thereby shaping the system's overall efficiency and output. Therefore, the accuracy and performance of the entire model heavily depend on the capabilities of the underlying LLM. As advancements in LLMs continue, their enhanced computational power and refined algorithms can significantly boost the performance of multi-agent systems. Moreover, the modular architecture of our approach allows for the integration of different LLMs tailored to the specific needs of individual agents. The integration of high-performance, open-source foundation models such as Llama 3 and Mistral/Mixtral, alongside specialized smaller models like Phi-3 or fine-tuned models in scientific subjects \cite{luu2024bioinspiredllm}, opens up exciting possibilities for further enhancement \cite{luu2024learning}. As these foundation models continue to advance in capability, the potential for our multi-agent systems to improve becomes even more pronounced. This adaptability not only ensures that our systems stay at the cutting edge of technological advancements but also signifies that the effectiveness and efficiency of AtomAgents will invariably increase. By continuously incorporating more capable LLMs, our systems can achieve deeper insights and more precise predictions, leading to accelerated innovation in materials design and discovery.

\section{Materials and Methods}\label{sec:methods}

\subsection*{Screw dislocation generation}
To model the screw dislocation, a periodic array of dislocations (PAD) configuration is used (e.g. ref. \cite{bacon2009dislocation}) with periodic boundary conditions along the dislocation glide direction z||[$11\overline{2}$] and dislocation line direction x||[111], and free boundaries along the glide plane normal direction y||[$\overline{1}10$]. Atomic positions are relaxed by using a combination of the FIRE algorithm \cite{bitzek2006structural} and relaxation of the cell dimensions until the convergence is achieved-the norm of the force vector fell below 10$^{-6}$ eV/ $\text{\AA}$ and stresses $\sigma_{xx}$ , $\sigma_{xy}$ , and $\sigma_{yy}$ fell below 0.1 MPa.

\subsection*{NEB simulations}\label{sec:NEB_simulation}
To compute the minimum energy path between initial and final screw dislocation configurations, nudged elastic band (NEB) \cite{mills1998nudged, henkelman2000climbing, henkelman2000improved} computations are performed as implemented in LAMMPS. To perform NEB simulations, first the initial and final screw dislocation configurations were created inside the material using the PAD method. Then 
NEB simulations are performed using the desired number of replicas. An
initial path of intermediate configurations (replicas) is constructed
by linearly interpolating the atomic positions between the relaxed
initial and final states. The NEB inter-replica spring constant is set
to $10^{-2}$ eV/$\text{\AA}^{2}$ and convergence is assumed when the maximum of
the force acting on all of the atoms across all replicas is less than
10$^{-3}$ eV/$\text{\AA}$. 
\subsection*{Surface energy calculations}
The surface energies along a given plane are computed as follows. First,  system with periodic boundary conditions in all directions is created and relaxed and the energy $E^{bulk}$ is computed. Then, the supercell is extended along the desired direction and then the  system is relaxed and the energy is computed $E^{surf}$. The surface energy is then computed as
\begin{equation*}
\gamma_{s}=\frac{E^{surf}-E^{bulk}}{2 A}
\end{equation*}
where $A$ is the area of the surface plane. 

\subsection{Critical fracture toughness}\label{sec:fracture_toughness}
The mode I anisotopic critical fracture toughness is given by
\begin{equation*}
    K_{Ic} = \sqrt{\frac{2\gamma_s}{\lambda_{22}}}
\end{equation*}
where $\gamma_{s}$ is the surface energy and $\lambda_{22}$ is the component of the the Stroh energy tensor \cite{andric2018atomistic, stroh1958dislocations}.
\subsection*{Agent design}
We design AI agents using the state-of-the-art all-purpose LLM GPT-4 family models. The dynamic multi-agent collaboration is implemented in the AutoGen framework \cite{wu2023autogen}, an open-source ecosystem for agent-based AI modeling.  In our multi-agent system, the user, admin, and executor agents are constructed using UserProxyAgent class from Autogen, and assistant and reviewer agents in the ``Knowledge retrieval" tool are crated using RetrieveAssistantAgent class. Moreover, the ``Plot analyzer" agent is created via the MultimodelConversableAgent class. The remaining agents are created via AssistantAgent class from Autogen; and the group chat manager is created using GroupChatManager class. Each agent is assigned a role through a profile description, included as \textit{system\_message} at their creation.   Details of the implementation can be identified in the source code shared via GitHub. 

\subsection*{Function and tool design}
All the tools implemented in this work are defined as Python functions. Each function is characterized by a name, a description, and input properties with a description.

\subsection*{Conflict of interest}
The author declares no conflict of interest.

\subsection*{Data and code availability}
All data and codes are available on GitHub at \url{https://github.com/lamm-mit/AtomAgents}. Alternatively, they will be provided by the corresponding author based on reasonable request.  
\\
\\

\subsection*{Supplementary Materials}
Additional materials are provided as Supplementary Materials. 

\section*{Acknowledgements}
We acknowledge support from USDA (2021-69012-35978), DOE-SERDP (WP22-S1-3475), ARO (79058LSCSB, W911NF-22-2-0213 and W911NF2120130) as well as the MIT-IBM Watson AI Lab, MIT’s Generative AI Initiative, and Google. Additional support from NIH (U01EB014976 and R01AR077793) is acknowledged. AG gratefully acknowledges the financial support from the Swiss National Science Foundation (project \#P500PT\_214448).

%Bibliography
\bibliographystyle{unsrt}  
\bibliography{library}

\begin{thebibliography}{100}
\bibitem{}
Significant-gravitas/autogpt: Autogpt is the vision of accessible ai for everyone, to use and to build on. our mission is to provide the tools, so that you can focus on what matters.
\end{thebibliography}

\end{document}